\documentclass[times,twocolumn,final,authoryear]{elsarticle}

\usepackage{ycviu}
\usepackage{framed,multirow}
\usepackage{lineno,hyperref}
\usepackage{amsthm}
\usepackage{graphicx}
\usepackage{algorithm}
\usepackage{caption}
\usepackage{subcaption}
\usepackage{algpseudocode}
\usepackage{amsmath,amssymb}
\usepackage{color}
\usepackage{tikz}
\usepackage{pgfplots}
\usepackage{pgf-umlsd}
\usepackage{ifthen}
\usepackage{amssymb}
\usepackage{latexsym}
\usepackage{url}
\usepackage{xcolor}
\definecolor{newcolor}{rgb}{.8,.349,.1}

\journal{Computer Vision and Image Understanding}

\begin{document}
\thispagestyle{empty}

\ifpreprint
  \setcounter{page}{1}
\else
  \setcounter{page}{1}
\fi
\begin{frontmatter}

\title{Learn to synthesize and synthesize to learn}

\author[1]{Behzad \snm{Bozorgtabar}\corref{cor1}} 
\cortext[cor1]{Corresponding author: 
  Tel.: +41-216934663;}
\ead{behzad.bozorgtabar@epfl.ch}
\author[1]{Mohammad Saeed \snm{Rad}}
\author[2]{Haz{\i}m Kemal Ekenel}
\author[1,3]{Jean-Philippe  \snm{Thiran}}

\address[1]{Signal Processing Laboratory (LT55), Ecole Polytechnique Féderale de Lausanne (EPFL-STI-IEL-LT55), Station 11, \\ 1015 Lausanne, Switzerland}
\address[2]{Istanbul Technical University, Istanbul, Turkey}
\address[3]{Department of Radiology, University Hospital Center (CHUV), University of Lausanne (UNIL), Lausanne, Switzerland}


\begin{abstract}
Attribute guided face image synthesis aims to manipulate attributes on a face image. Most existing methods for image-to-image translation can either perform a fixed translation between any two image domains using a single attribute or require training data with the attributes of interest for each subject. Therefore, these methods could only train one specific model for each pair of image domains, which limits  their ability in dealing with more than two domains. Another disadvantage of these methods is that they often suffer from the common problem of mode collapse that degrades the quality of the generated images. To overcome these shortcomings, we propose attribute guided face image generation method using a single model, which is capable to synthesize multiple photo-realistic face images conditioned on the attributes of interest. In addition, we adopt the proposed model to increase the realism of the simulated face images while preserving the face characteristics. Compared to existing models, synthetic face images generated by our method present a good photorealistic quality on several face datasets. Finally, we demonstrate that generated facial images can be used for synthetic data augmentation, and improve the performance of the classifier used for facial expression recognition. 
\end{abstract}


\end{frontmatter}

\section{Introduction}
\label{sec:intro}
\begin{figure*}[h]
\centering
\includegraphics[height=10.5cm, width=16.5cm]{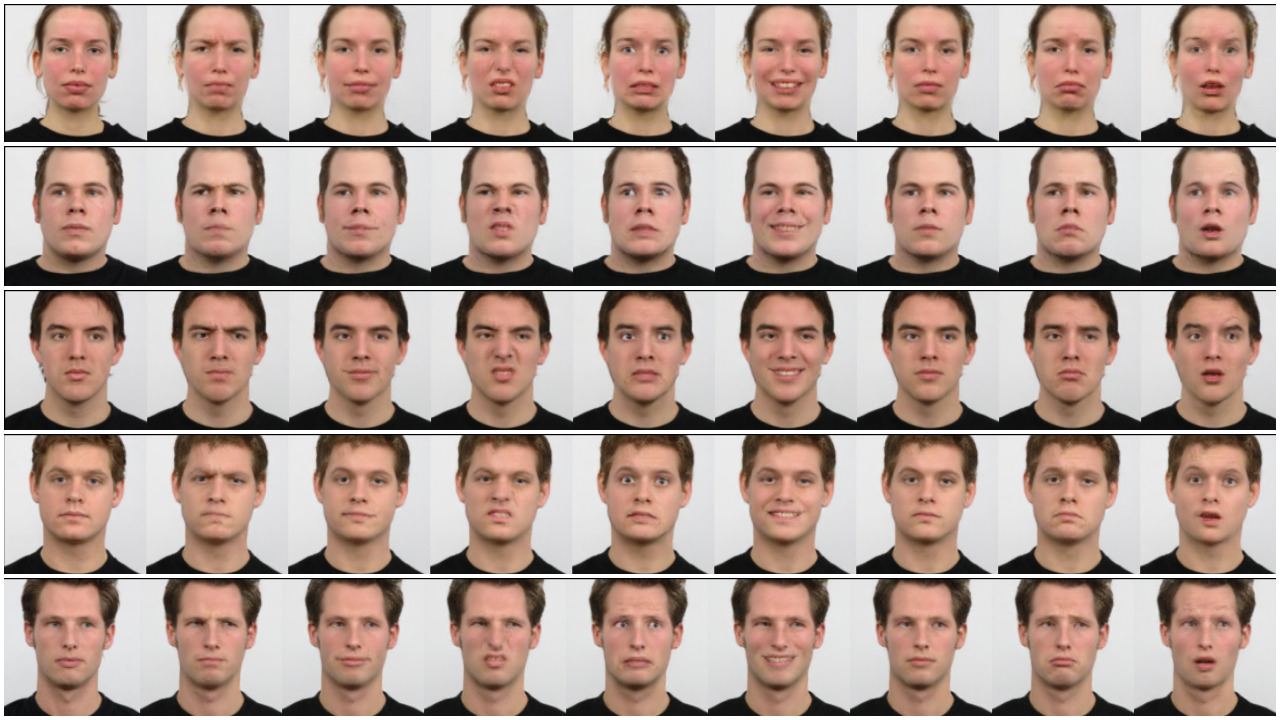}
  \caption{Attribute guided facial image generation using LSSL on the Radboud Faces Database (RaFD) \cite{langner2010presentation}. The input \textit{neutral} faces are fed into our model to exhibit specified attribute. \textbf{Left to right}: input \textit{neutral} face and seven different attributes including \textit{angry}, \textit{contemptuous}, \textit{disgusted}, \textit{fearful}, \textit{happiness}, \textit{neutral}, \textit{sadness} and \textit{surprised}, respectively.}
\label{fig:1}
\end{figure*}

In this work, we are interested in the problem of synthesizing realistic faces by controlling the facial attributes of interest (e.g. expression, pose, lighting condition) without affecting the identity properties (see Fig. \ref{fig:1}). In addition, this paper investigates learning from synthetic facial images for improving expression recognition accuracy. Synthesizing photo-realistic facial images has applications in human-computer interactions, facial animation and more importantly in facial identity or expression recognition. However, this task is challenging since image-to-image translation is ill-defined problem and it is difficult to collect images of varying attributes for each subject (e.g. images of different facial expressions for the same subject). The most notable solution is the incredible breakthroughs in generative models. In particular, Generative Adversarial Network (GAN) \cite{goodfellow2014generative} variants have achieved state-of-the-art results for the image-to-image translation task. These GAN models could be trained in both with paired training data \cite{isola2017image} and unpaired training data \cite{kim2017learning,zhu2017unpaired}. Most existing GAN models \cite{shen2017learning,zhu2017unpaired} are proposed to synthesize images of a single attribute, which make their training inefficient in the case of having multiple attributes, since for each attribute a separate model is needed. In addition, GAN based approaches are often fragile in the common problem of mode collapse that degrades the quality of the generated images. To overcome these challenges, our objective is to use a single model to synthesize multiple photo-realistic images from the same input image with varying attributes simultaneously. Our proposed model, namely Lean to Synthesize and Synthesize to Learn (LSSL) is based on encoder-decoder structure, using the image latent representation, where we model the shared latent representation across image domains. Therefore, during inference step, by changing input face attributes, we can generate plausible face images owing attribute of interest. We introduce bidirectional learning for the latent representation, which we have found this loss term to prevent generator mode collapse. Moreover, we propose to use an additional face parsing loss to generate high-quality face images.

Our paper makes the following contributions:

\begin{enumerate}
\item This paper investigates domain adaptation using simulated face images for improving expression recognition accuracy. We show that how the proposed approach can be used to generate photo-realistic frontal facial images using synthetic face image and unlabeled real face images as the input. We compared our results with SimGAN method \cite{shrivastava2017learning} in terms of expression recognition accuracy to see improvement in the realism of frontal faces. The source code is available at {\url{https://github.com/CreativePapers/Learn-to-Synthesize-and-Synthesize-to-Learn}}.

\item We show that use of our method leads to realistic generated images that contribute to improve the performance of expression recognition accuracy despite having small number of real training images. Further, compared to other variants of GAN models \cite{zhu2017unpaired,perarnau2016invertible,choi2018stargan}, we show that a better performance can be attained through a proposed method to focus on the data augmentation process;

\item Unlike most of existing GAN based methods \cite{perarnau2016invertible}, which are trained with a large number of labeled and matching image pairs, the proposed method is adopted for unpaired image-to-image translation. As a matter of fact, the proposed method transfers the learnt characteristics between different classes; 

\item The proposed method is capable of learning image-to-image translation among multiple domains using a single model. We introduce a bidirectional learning for the image latent representation to additionally enforce latent representation to capture shared features of different attribute categories and to prevent generator mode collapse. By doing so, we synthesize face photos with a desired attribute and translate an input image into another domain image\footnote{We denote \textit{domain} as a set of images owning the same attribute value.}. Besides, we present face parsing loss and identity loss that help to preserve the face image local details and identity.
\end{enumerate}

\section{Related work}
\label{sec:related}
Recently, GAN based models \cite{goodfellow2014generative} have achieved impressive results in many image synthesis applications, including image super-resolution \cite{ledig2017photo}, image-to-image translation (pix2pix) \cite{isola2017image} and CycleGAN \cite{zhu2017unpaired}. We summarize contributions of few important related works in below:

\paragraph{Applications of GANs to Face Generation}
\cite{taigman2016unsupervised} proposed a domain transfer network to tackle the problem of emoji generation for a given facial image. \cite{lu2018attribute} proposed attribute-guided face generation to translate low-resolution face images to high-resolution face images. \cite{huang2017beyond} proposed a Two-Pathway Generative Adversarial Network (TP-GAN) for photorealistic face synthesis by simultaneously considering local face details and global structures.

\paragraph{Image-to-Image Translation Using GANs}
Many of existing image-to-image translation methods e.g. \cite{isola2017image, shrivastava2017learning} formulated GANs in the supervised setting, where example image pairs are available. However, collecting paired training data can be difficult. On the other side, there are other GAN based methods, which do not require matching pairs of samples. For example, CycleGAN \cite{zhu2017unpaired} is capable to learn transformations from source to target domain without one-to-one mapping between two domain's training data. \cite{li2016deep} proposed a Deep convolutional network model for Identity-Aware Transfer (DIAT) of the facial attributes. However, these GAN based methods could only train one specific model for each pair of image domains. Unlike the aforementioned approaches, we use a single model to learn to synthesize multiple photo-realistic images, each having specific attribute. More recently, IcGAN \cite{perarnau2016invertible} and StarGAN \cite{choi2018stargan} proposed image editing using AC-GAN \cite{odena2017conditional} with conditional information. However, we use domain adaptation by adding the realism to the simulated faces and there is no such a solution in these methods. Similar to \cite{perarnau2016invertible}, Fader Networks \cite{lample2017fader} proposed image synthesis model without needing to apply a GAN to the decoder output. However, these methods impose constraints on image latent space to enforce it to be independent from the attributes of interest, which may result in loss of information in generating attribute guided images.

\paragraph{GANs for Facial Frontalization and Expression Transfer}
\cite{zhang2018joint} proposed a method by disentangling the attributes (expression and pose) for simultaneous pose-invariant facial expression recognition and face images synthesis. Instead, we seek to learn attribute-invariant information in the latent space by imposing auxiliary classifier to classify the generated images. \cite{qiao2018geometry} proposed a Geometry-Contrastive Generative Adversarial Network (GC-GAN) for transferring continuous emotions across different subjects. However, this requires a training data with expression information, which may be expensive to obtain. Alternatively, our self-supervised approach automatically learns the required factors of variation by transferring the learnt characteristics between different emotion classes. \cite{zhu2018emotion} investigated GANs for data augmentation for the task of emotion classification. \cite{lai2018emotion} proposed a multi-task GAN-based network that learns to synthesize the frontal face images from profile face images. However, they require paired training data of frontal and profile faces. Instead, we seek to add realism to the synthetic frontal face images without requiring real frontal face images during training. Our method could produce synthesis faces using synthetic frontal faces and real faces with arbitrary poses as input.

\section{Methods}
\label{sec:approach}
We first introduce our proposed multi-domain image-to-image translation model in Section \ref{subsec:attribute}. Then, we explain learning from simulated data by adding realism to simulated face images in Section \ref{subsec:posenormalization}. Finally, we discuss our implementation details and experimental results in Section \ref{implementation} and Section \ref{subsec:experimentalresults}, respectively.

\subsection{Learn to Synthesize}
\label{subsec:attribute}
\begin{figure}[h]
\centering
\begin{subfigure}[b]{.48\linewidth}
\includegraphics[width=\linewidth, height=1.1\linewidth]{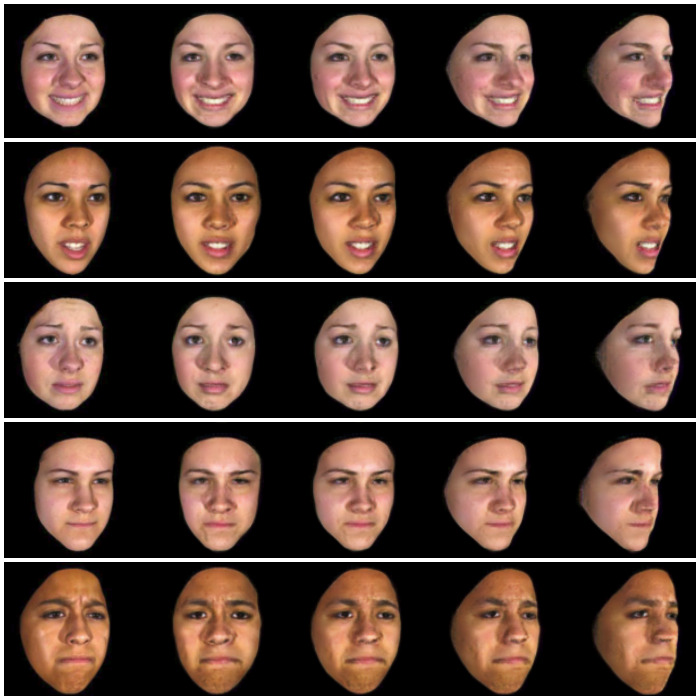}
\caption{}\label{pose}
\end{subfigure}
\begin{subfigure}[b]{.48\linewidth}
\includegraphics[width=\linewidth, height=1.1\linewidth]{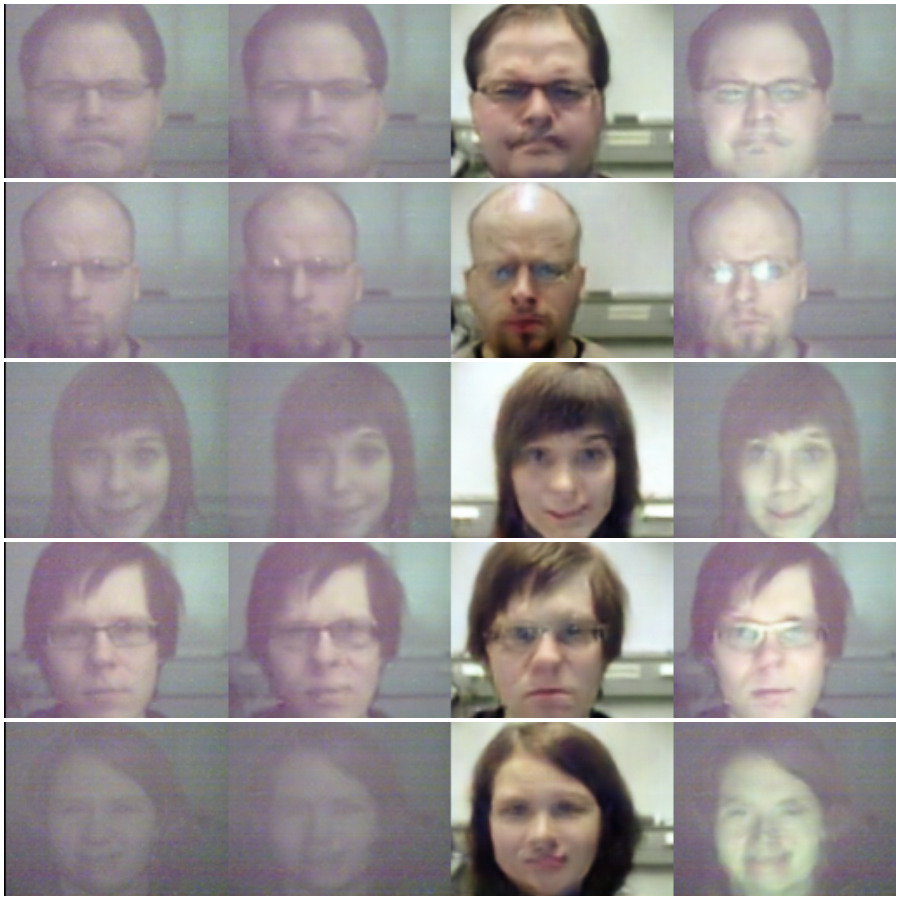}
\caption{}\label{lighting}
\end{subfigure}
\caption{Examples of facial attribute transfer. (a) Generating images with varying poses ranging from 0 to 45 degrees (yaw angle) in 15 degrees steps. (b) Generating face images with three different lighting conditions using face image with normal illumination as input: normal illumination (reconstruction), weak illumination and dark illumination, respectively.}
\label{fig:2}
\end{figure}

Let $\mathcal{X}$ and $\mathcal{S}$ denote original image and side conditional image domains, respectively and $\mathcal{Y}$ set of possible facial attributes, where we consider attributes including facial expression, head pose and lighting (see Fig. \ref{fig:2}). As the training set, we have $m$ triple inputs $\left (x_{i}\in \mathcal{X}, s_{i}\in \mathcal{S}, y_{i}\in \mathcal{Y} \right )$, where $x_{i}$ and $y_{i}$ are the $i^{th}$ input face image and binary attribute, respectively and $s_{i}$ represents the $i^{th}$ conditional side image as additional information to guide photo-realistic face synthesis. Then, for any categorical attribute vector $y$ from the set of possible facial attributes $\mathcal{Y}$, the objective is to train a model that will generate photo-realistic version (${x}'$ or ${s}'$) of the inputs ($x$ and $s$) from image domains $\mathcal{X}$ and $\mathcal{S}$ with desired attributes $y$.

\begin{figure*}
\centering
   \begin{subfigure}[b]{0.90\textwidth}
   \includegraphics[width=1\textwidth]{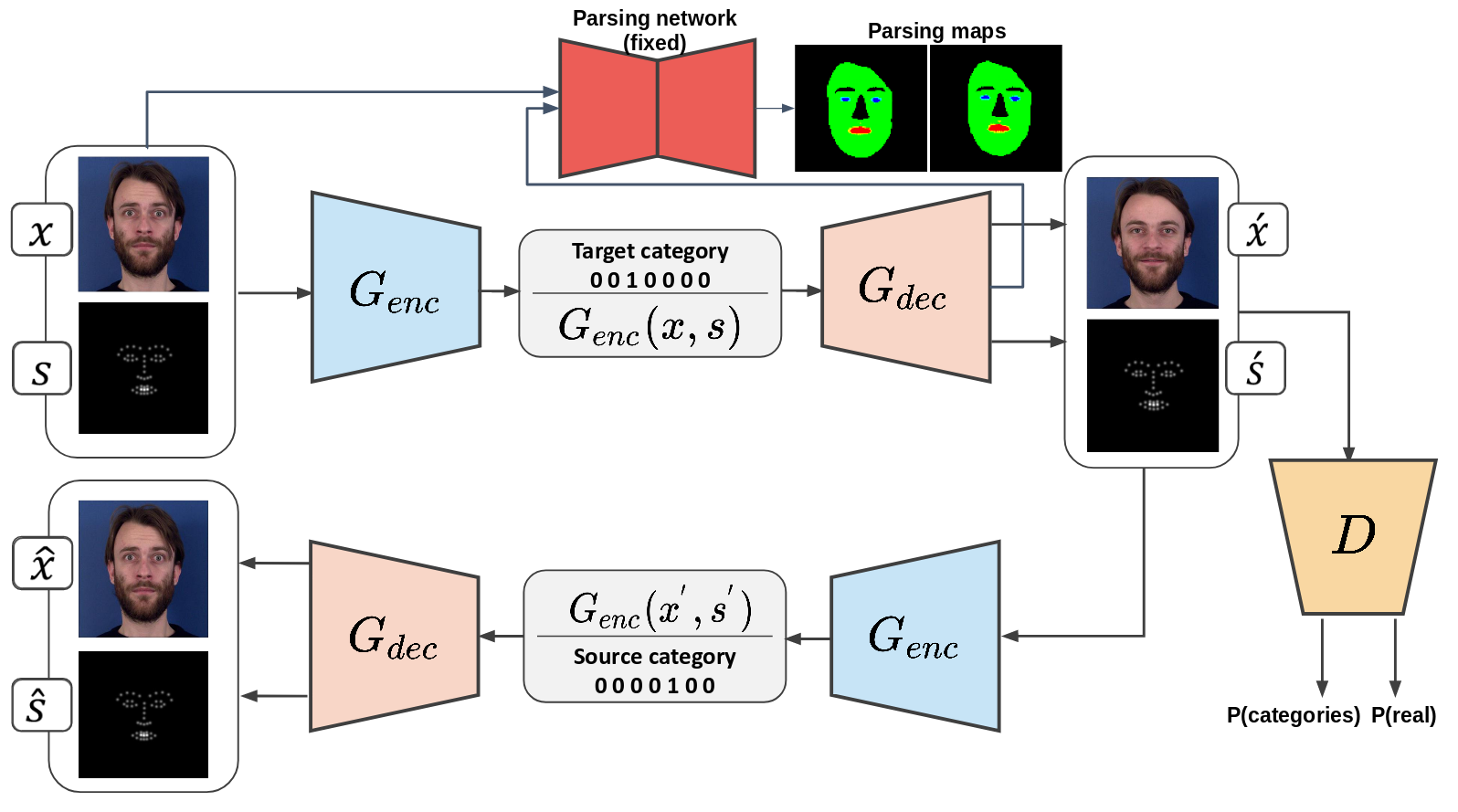}
   \caption{}
   \label{fig:3_1} 
\end{subfigure}
\begin{subfigure}[b]{0.90\textwidth}
   \includegraphics[width=1\textwidth]{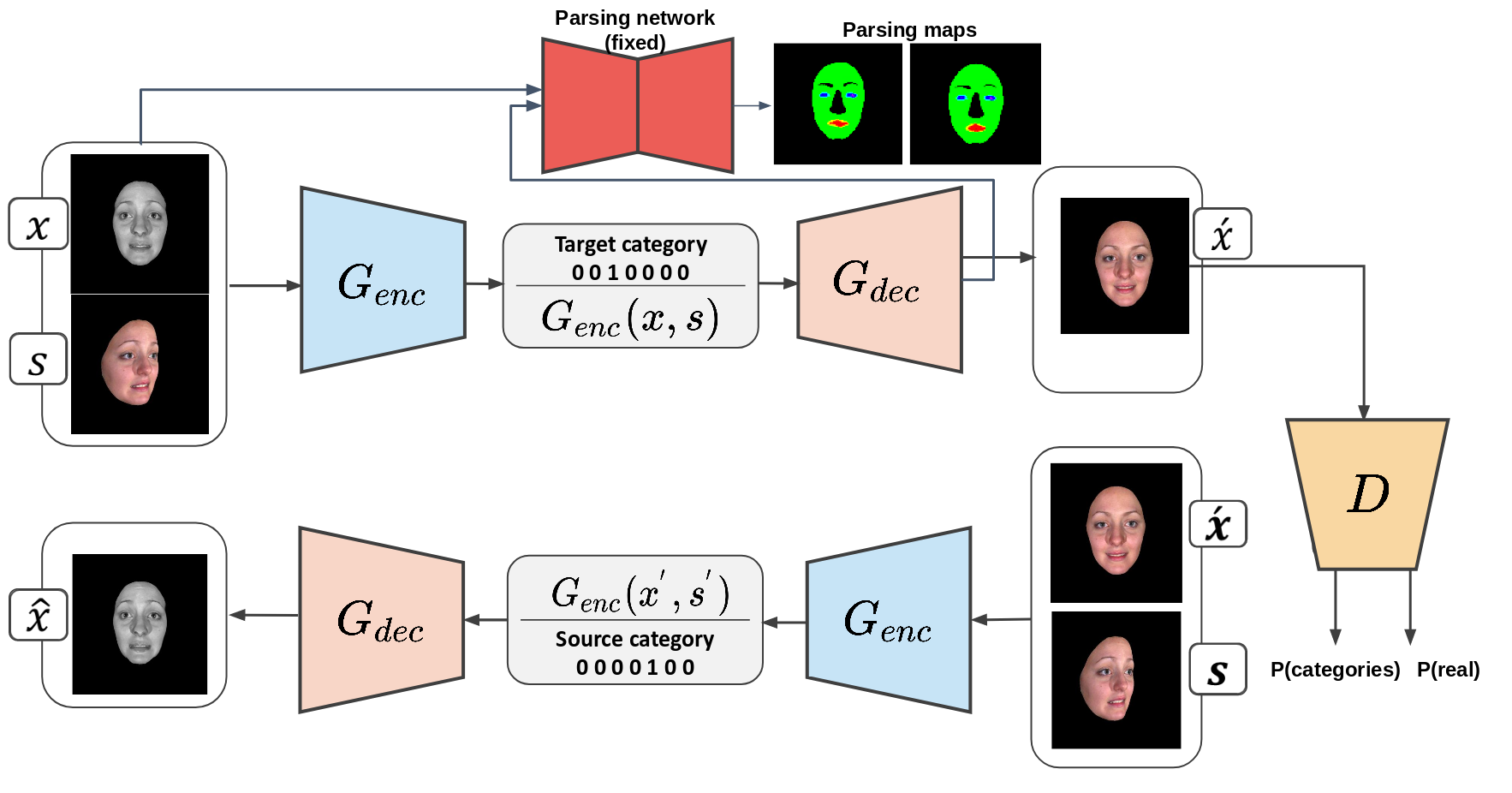}
   \caption{}
   \label{fig:3_2}
\end{subfigure}
\caption{Overview of our proposed LSSL method. (a) Attribute-guided face image synthesis model, consisting of three networks, an encoder-decoder generator $G$, a discriminator $D$ and face parsing network $P$. A discriminator's job is to discriminate the realism of synthetic pair images and to guarantee correct attribute classification on the generated face images. (b) Pose normalization model, which takes a synthetic face image and unlabeled real image as input and generates photo-realistic version of the frontal face through identical networks used in (a). The only difference is that a discriminator takes only one single image as input.}
\end{figure*}

Our model is based on the encoder-decoder architecture with domain adversarial training. As the input to our expression synthesis method (see Fig. \ref{fig:3_1}), we propose to incorporate individual-specific facial shape model as the side conditional information $s$ in addition to the original input image $x$. The shape model can be extracted from the configuration of the facial landmarks, where the facial geometry varies with different individuals. Our goal is then to train a single generator $G$ with encoder $G_{enc}$ -- decoder $G_{dec}$ networks to translate the input pair $\left ( x,s \right )$ from source domains into their corresponding output images $\left ( {x}',{s}' \right )$ in the target domain conditioned on the target domain attribute $y$ and the inputs latent representation $G_{enc}\left ( x,s \right )$, $G_{dec}\left ( G_{enc}\left ( x,s \right ),y \right )\rightarrow {x}',{s}'$. 
The encoder $G_{enc}:\left ( \mathcal{X}^{source}, \mathcal{S}^{source} \right )\rightarrow \mathbb{R}^{n\times \frac{h}{16}\times \frac{w}{16}}$ is a fully convolutional neural network with parameters $\theta _{enc}$ that encodes the input images into a low-dimensional feature space $G_{enc}\left ( x,s \right )$, where $n, h, w$ are the number of the feature channels and the input images dimensions, respectively. The decoder $G_{dec}:\left ( \mathbb{R}^{n\times \frac{h}{16}\times \frac{w}{16}},\mathcal{Y} \right )\rightarrow \mathcal{X}^{target}, \mathcal{S}^{target}$ is the sub-pixel \cite{shi2016real} convolutional neural network with parameters $\theta _{dec}$ that produce realistic images with target domain attribute $y$ and given the latent representation $G_{enc}\left ( x,s \right )$. The precise architectures of the neural networks are described in Section \ref{networkarchitechure}. During training, we randomly use a set of target domain attributes $y$ to make the generator more flexible in synthesizing images. In the following, we introduce the objectives for the proposed model optimization.

\paragraph{GAN Loss} We introduce a model that
discovers cross-domain image translation with GANs. Moreover, at the inference time, we should be able to generate diverse facial images by only changing attribute of interest. By doing so, we seek to learn attribute-invariant information in the latent space representing the shared features of the images sampled for different attributes. It means if the original and target domains are semantically similar (e.g. facial images of different expressions), we expect the common features across domains to be captured by the same latent representation. Then, the decoder must use the target attribute to perform image-to-image translation from the original domain to the target domain. However, this learning process is unsupervised as for each training image from the source domain, its counterpart image in the target domain with attribute $y$ is unknown. Therefore, we propose to train an additional neural network called the discriminator $D$ (with the parameters $\theta _{dis}$) using an adversarial formulation to not only distinguish between real and fake generated images, but also to classify the image to its corresponding attribute categories. We use Wasserstein GAN \cite{gulrajani2017improved} objective with a gradient penalty loss $\mathcal{L}_{gp}$ \cite{arjovsky2017wasserstein} formulated as below:
\begin{equation} \label{eq1}
\begin{split}
\mathcal{L} _{GAN}=\mathbb{E}_{x,s}\left [ D_{src}\left ( x,s  \right ) \right ]-\mathbb{E}_{x,s,y}\left [ D_{src}\left ( G_{dec}\left ( G_{enc}\left ( x,s \right ),y \right ) \right ) \right ]\\
-\lambda_{gp} \thinspace \mathcal{L}_{gp}\left ( D_{src} \right ), 
\end{split}
\end{equation}
The term $D_{src}\left ( \cdot  \right )$ denotes a probability distribution over image sources given by $D$. The hyper-parameter
$\lambda_{gp}$ is used to balance the GAN objective with the gradient penalty. A generator (encoder-decoder networks) used in our model has to play two roles: learning the attribute invariance representation for the input images and is trained to maximally fool the discriminator in a \textit{min-max} game. On the other hand, the discriminator simultaneously seeks to identify the fake examples for each attribute.

\paragraph{Attribute Classification Loss} We deploy a classifier by returning additional output from the discriminator to perform an auxiliary task of classifying the synthesized and real facial images into their respective attribute categories. An attribute classification loss of real images $\mathcal{L}_{cls_{r}}$ to optimize the discriminator parameters $\theta _{dis}$ is defined as follow:
\begin{equation} \label{eq2}
\begin{split}
\min\limits_{\theta _{dis}}\mathcal{L}_{cls_{r}}& =\mathbb{E}_{x,s,{y}'}\left [ \ell_{r}\left ( x,s,{y}' \right ) \right ],\\
\ell_{r}\left ( x,s,{y}' \right )& =\sum_{i=1}^{m}-{y_{i}}'\log D_{cls}\left ( x,s \right )-\left ( 1-{y_{i}}' \right )\log\left ( 1-D_{cls}\left ( x,s \right ) \right ),
\end{split}
\end{equation}
Here, ${y}'$ denotes original attributes categories for the real images. $\ell_{r}$ is the summation of binary cross-entropy losses of all attributes. Besides, an attribute classification loss of fake images $\mathcal{L}_{cls_{f}}$ used to optimize the generator parameters $\left ( \theta _{enc},\theta _{dec} \right )$, formulated as follow:
\begin{equation} \label{eq3}
\begin{split}
\min\limits_{\theta _{enc},\theta _{dec}}\mathcal{L}_{cls_{f}} =\mathbb{E}_{x,s,{y}'}\left [ \ell_{f}\left ( {x}',{s}',y \right ) \right ],\\
\ell_{f}\left ( {x}',{s}',y \right ) =\sum_{i=1}^{m}-y_{i}\log D_{cls}\left ( {x}',{s}' \right )\\
-\left ( 1-y_{i} \right )\log\left ( 1-D_{cls}\left ( {x}',{s}' \right ) \right ),
\end{split}
\end{equation}
where ${x}'$ and ${s}'$ are the generated images and auxiliary outputs, which should correctly own the target domain attributes $y$. $\ell_{f}$ denotes summing up the cross-entropy losses of all fake images.

\begin{figure}[h]
\centering
\includegraphics[height=6.5cm, width=9cm]{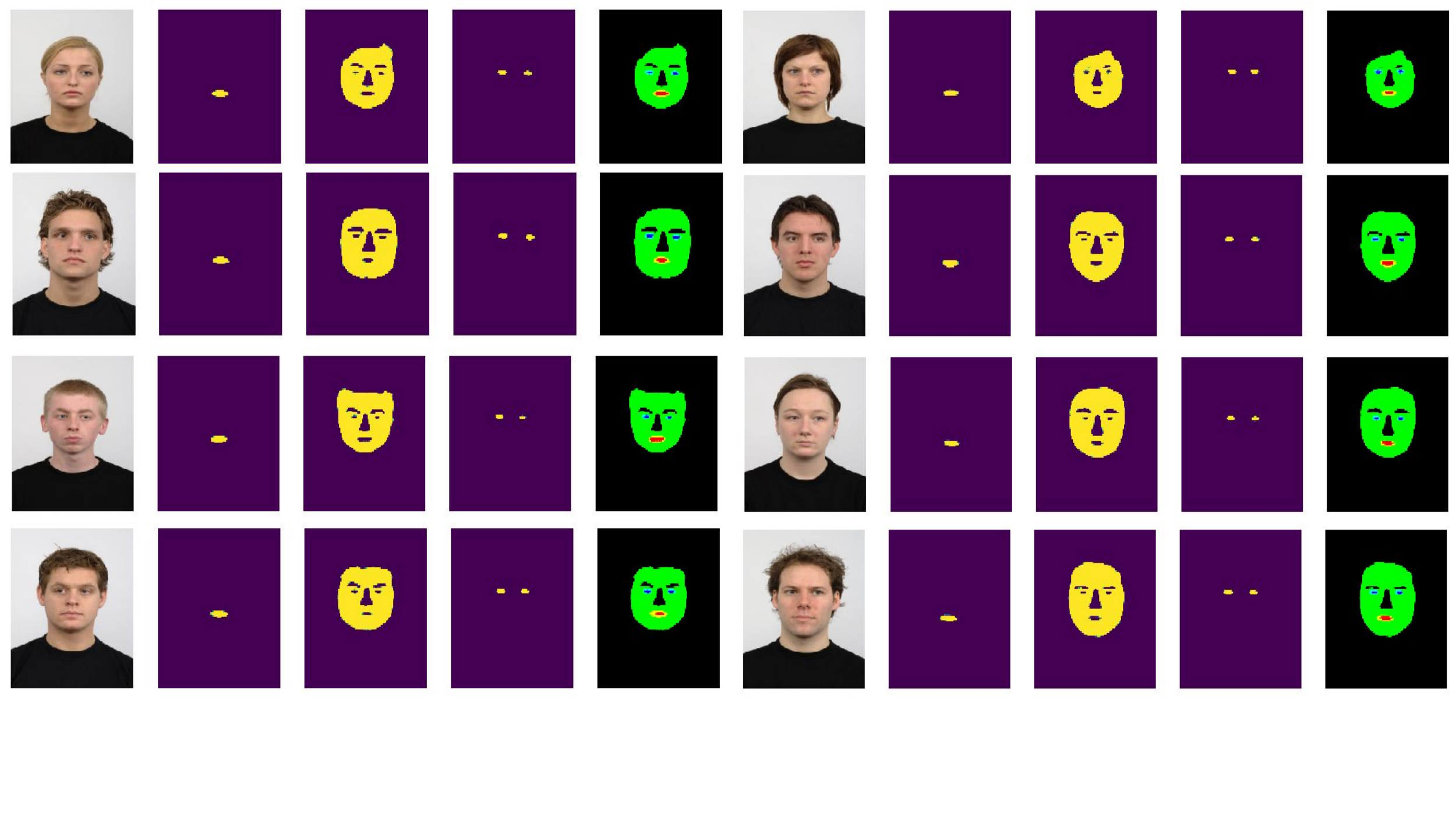}
  \caption{Face parsing maps on the RaFD dataset. \textbf{Left to right}: input \textit{neutral} face and parsing maps for its constituent facial parts containing lips (second column), face skin (third column), eyes (fourth column) and color visualization generated by all three category parsing maps (last column), respectively. }
\label{fig:4}
\end{figure}

\paragraph{Identity Loss} Using the identity loss, we aim to preserve the attribute-excluding facial image details such as facial identity before and after image translation. By doing so, we use a pixel-wise $l_{1}$ loss to enforce the details consistency of the face original domain and suppress the face blurriness:

\begin{equation} \label{eq4}
\begin{split}
\mathcal{L}_{id} =
\mathbb{E}_{x,s,{y}'}\left [\left \|  G_{dec}\left ( G_{enc}\left ( x,s \right ),y \right )-x\right \|_{1}  \right ],
\end{split}
\end{equation}
\paragraph{Face Parsing Loss}
The face important components (e.g., lips and eyes) are typically small and cannot be well reconstructed by solely minimizing the identity loss on the whole face image. Therefore, we use a face parsing loss to further improve the harmony of the synthetic faces. As our face parsing network, we use U-Net \cite{ronneberger2015u} trained on the Helen dataset \cite{le2012interactive}, which has ground truth face semantic labels, for training parsing network. Instead of utilizing all semantic labels, we use three key face components (lips, eyes and face skin). Once the network is trained, it remains fixed in our framework. The parsing loss is back-propagated to the generator to further regularize generator. Fig. \ref{fig:4} shows some parsing results on the RaFD dataset \cite{langner2010presentation}. 

\begin{equation} \label{eq5}
\begin{split}
\mathcal{L}_{p} =\mathbb{E}_{x,s,{y}'}\left [ A_{p}\left ( P\left ( x \right )-P\left ( {x}' \right ) \right ) \right ],
\end{split}
\end{equation}

where $A_{p}\left ( \cdot ,\cdot  \right )$ denotes a function to compute pixel-wise softmax loss and $P\left ( \cdot  \right )$ is the face parsing network.
\paragraph{Bidirectional Loss} Using GAN loss alone usually leads to mode collapse, generating identical labels regardless of the input face photo. This problem has been observed in various applications of conditional GANs \cite{isola2017image,dosovitskiy2016generating} and to our knowledge, there is still no proper approach to deal with this issue. To address this problem, we show that using the trained generator, images of different domains can be translated bidirectionally. We decompose this objective into two terms: a bidirectional loss for the image latent representation, and a bidirectional loss between synthesized images and original input images, respectively. This objective is formulated using $l_{1}$ loss as follow:
\begin{equation} \label{eq6}
\begin{split}
\mathcal{L} _{bi} =\mathbb{E}_{x,s,{y}'}\left [\left \| x-\hat{x} \right \|_{1}+\left \| s-\hat{s} \right \|_{1}  \right ]+\\
\mathbb{E}_{x,s,y}\left [\left \| G_{enc}\left ( x,s \right )-G_{enc}\left ( {x}',{s}' \right ) \right \|_{1}  \right ], \\
{x}',{s}'=G_{dec}\left ( G_{enc}\left ( x,s \right ),y \right ), \\
\hat{x},\hat{s} =G_{dec}\left ( G_{enc}\left ( {x}',{s}' \right ),{y}' \right ),
\end{split}
\end{equation}

In the above equation, $\hat{x}$ and $\hat{s}$ denote the reconstructed original image and the side conditional image, respectively. Unlike \cite{zhu2017unpaired}, where only the cycle consistency losses are used at the image level, we additionally seek to minimize the reconstruction loss using latent representation.

\paragraph{Overall Objective} Finally, the generator $G$ is trained with a linear combination of five loss
terms: adversarial loss, attribute classification loss for the fake images, bidirectional loss, identity loss and face parsing loss. Meanwhile, the discriminator $D$ is optimized using an adversarial loss and attribute classification loss for the real images:

\begin{equation} \label{eq7}
\begin{split}
\mathcal{L}_{G}&=\mathcal{L}_{GAN}+\lambda _{bi}\mathcal{L}_{bi}+\lambda _{cls}\mathcal{L}_{cls_{f}}+\lambda _{id}\mathcal{L}_{id}+\lambda_{p}\mathcal{L}_{p},\\
\mathcal{L}_{D}&=-\mathcal{L}_{GAN}+\lambda _{cls}\mathcal{L}_{cls_{r}},
\end{split}
\end{equation}
where $\lambda _{bi}$, $\lambda _{p}$, $\lambda _{id}$ and $\lambda _{cls}$ are hyper-parameters, which tune the importance of bidirectional loss, face parsing loss, identity loss and attribute classification loss, respectively.

\subsection{Synthesize to Learn}
\label{subsec:posenormalization}
In an unconstrained face expression recognition, accuracy will drop significantly for large pose variations. The key solution would be using simulated faces rendered in frontal view. However, learning from synthetic face images can be problematic due to a distribution discrepancy between real and synthetic images. Here, our proposed model generates realistic face images given real profile face with arbitrary pose and a simulated face image as input (see Fig. \ref{fig:3_2}). We utilize a 3D Morphable Model using bilinear face model \cite{vlasic2005face} to construct a simulated frontal face image from multiple camera views. Here, the discriminator's role is to discriminate the realism of synthetic face images using unlabeled real profile face images as a conditional side information. In addition, using the same discriminator, we can generate face images exhibiting different expressions.

We compare the results of LSSL with SimGAN method \cite{shrivastava2017learning} on the BU-3DFE dataset \cite{yin20063d} to evaluate the realism of face images. SimGAN method \cite{shrivastava2017learning} considers learning from simulated and unsupervised images through adversarial training. However, SimGAN is devised for much simpler scenarios e.g., eye image refinement. In addition, categorical information was ignored in SimGAN, which limits the model generalization. In contrast, LSSL overcomes this issue by introducing attribute classification loss into objective function. For a fair comparison with SimGAN method, we add the attribute classification loss by modifying the SimGAN's discriminator, while keeping the rest of network unchanged. We achieve more visually pleasing results on test data compared to the SimGAN method (see Fig. \ref{fig:7}). 

\section{Implementation Details}
\label{implementation}
All networks are trained using Adam optimizer \cite{kingma2014adam} $\left ( \beta _{1}=0.5,\beta _{2}=0.999 \right )$ and with a base learning rate of $0.0001$. We linearly decay learning rate after the first 100 epochs. We use a simple data augmentation with only flipping the images horizontally. The input image size and the batch size are set to $128\times 128$ and 8 for all experiments, respectively. We update the discriminator five times for each generator (encoder-decoder) update. The hyper-parameters in Eq. \ref{eq7} and Eq. \ref{eq1} are set as: $\lambda _{bi}=10$ and $\lambda _{id}=10$, $\lambda _{p}=10$,  $\lambda_{gp}=10$ and $\lambda _{cls}=1$, respectively. The whole model is implemented using PyTorch on a single NVIDIA GeForce GTX 1080.

\subsection{Networks Architectures}
\label{networkarchitechure}
\begin{figure*}[h]
\centering
\includegraphics[height=10cm, width=17cm]{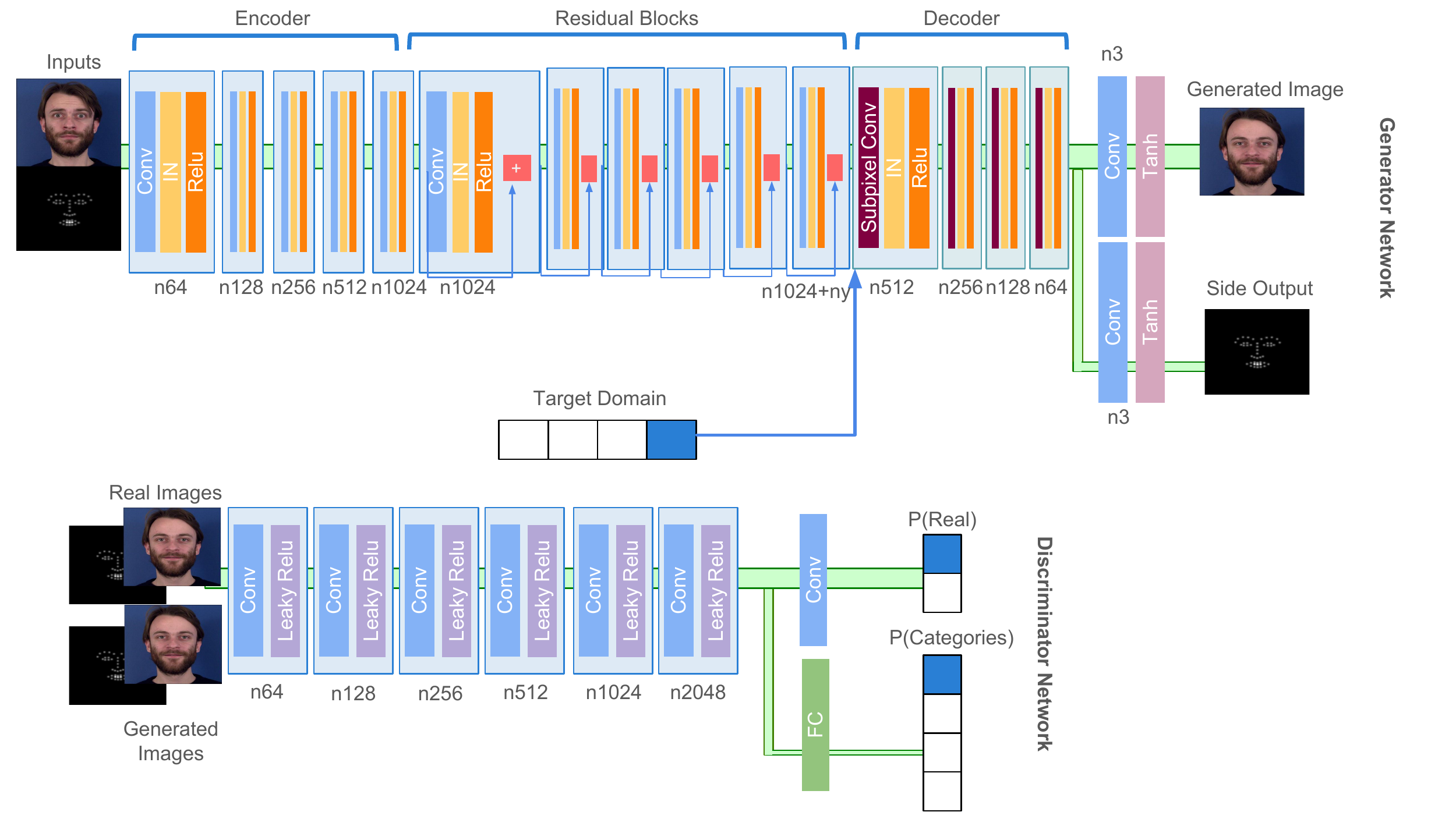}
  \caption{Architecture of generator (\textbf{top}) and discriminator network (\textbf{bottom}). $n_{y}$ denotes the the dimension of domain attributes. IN and $n$ and FC denote instance normalization, number of feature maps and fully connected layer, respectively.}
\label{fig:5}
\end{figure*}

For the discriminator, we use PatchGAN \cite{isola2017image} that penalizes structure at the scale of image patches. In addition, LSSI has the generator network composed of five convolutional layers with the stride size of two for downsampling, six residual blocks, and four transposed convolutional layers with the stride size of two for upsampling. We use sub-pixel convolution instead of transposed convolution followed by instance normalization \cite{ba2016layer}. For the face parsing network, we used the same net architecture as U-Net proposed in \cite{ronneberger2015u}, but our face parsing network consists of depthwise convolutional blocks proposed by MobileNets \cite{sandlerv2}. The network architecture of LSSL is shown in Fig. \ref{fig:5}.
\section{Experimental Results}
\label{subsec:experimentalresults}
In this section, we first propose to carry out comparison between our LSSL method and recent methods in image-to-image translation from a qualitative perspective, then we demonstrate the generality of our method (quantitative analysis) using different techniques for the face expression recognition.

\subsection{Datasets}
\textbf{Oulu-CASIA VIS \cite{zhao2011facial}}: This dataset contains 480 sequences (from 80 subjects) of six basic facial expressions under the visible (VIS) normal illumination conditions.  The sequences start from a neutral face and end with peak facial expression. This dataset is chosen due to high intra-class variations caused by the personal attributes. We conducted our experiments using subject-independent 10-fold cross-validation strategy.

\textbf{MUG \cite{aifanti2010mug}}: The MUG dataset contains image sequences of seven different facial expressions belonging to 86 subjects comprising 51 men and 35 women. The image sequences were captured with a resolution of $896\times 896$. We used image sequences of 52 subjects and the corresponding  annotation, which are available publicly via the internet. 

\textbf{BU-3DFE \cite{yin20063d}}: The Binghamton University 3D Facial Expression Database (BU-3DFE) \cite{yin20063d} contains 3D models from 100 subjects, 56 females and 44 males. The subjects show a neutral face as well as six basic facial expressions and at four different intensity levels. Following the setting in \cite{tariq2013maximum} and \cite{zhang2018joint}, we used an openGL based tool from the database creators to render multiple views from 3D models in seven pan angles $\left ( 0^{\circ},\pm 15^{\circ},\pm 30^{\circ},\pm 45^{\circ} \right )$.

\textbf{RaFD \cite{langner2010presentation} }: The Radboud Faces Database (RaFD) contains 4,824 images belonging to 67 participants. Each subject makes eight facial expressions.

\begin{figure}[h]
\centering
\includegraphics[height=7cm, width=8.6cm]{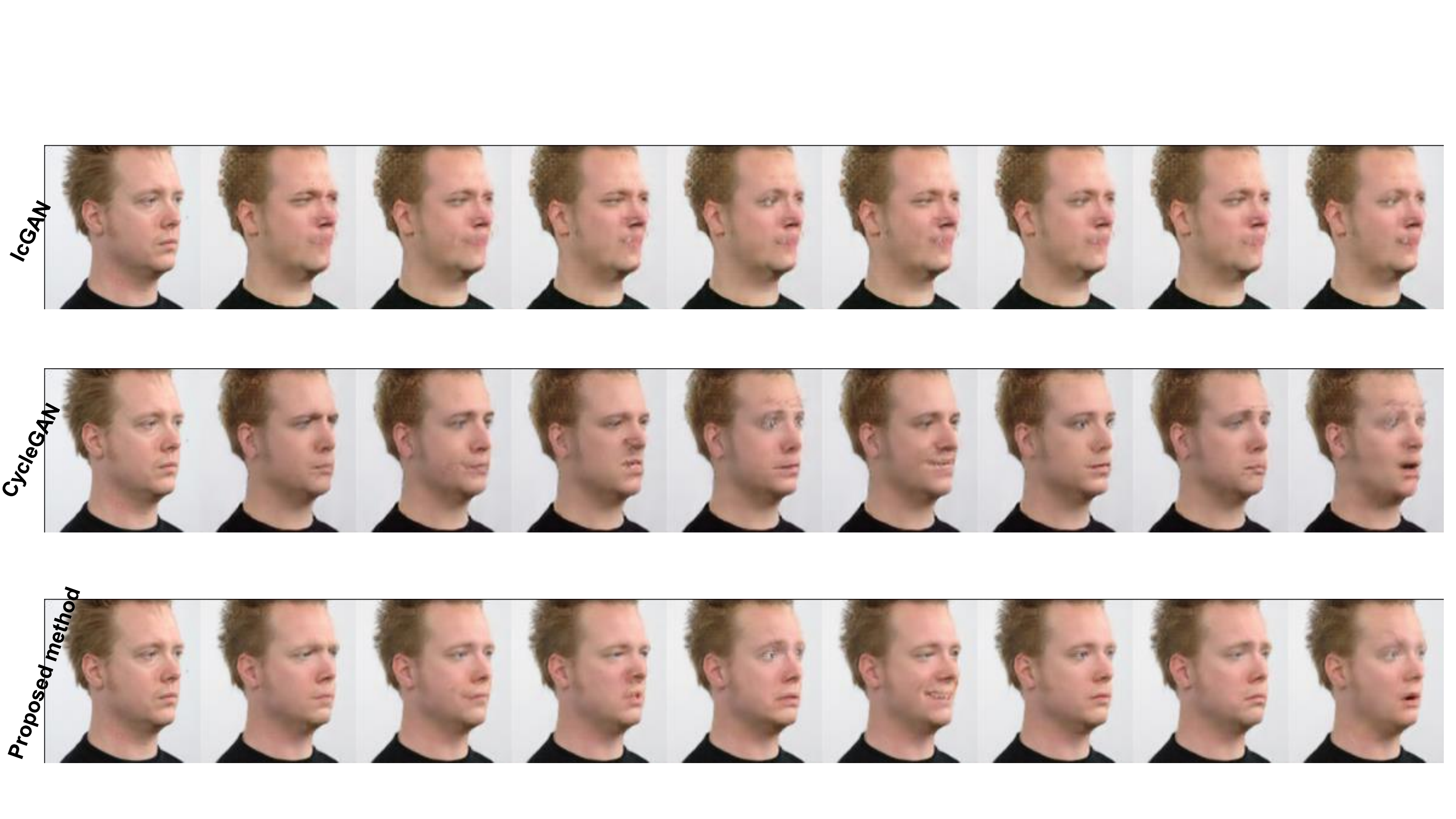}
  \caption{Facial attribute transfer results of LSSL compared with IcGAN \cite{perarnau2016invertible} and CycleGAN \cite{zhu2017unpaired}, respectively.}
\label{fig:6}
\end{figure}

\begin{figure}
\centering
\begin{subfigure}[a]{0.46\textwidth}
   \includegraphics[width=\linewidth]{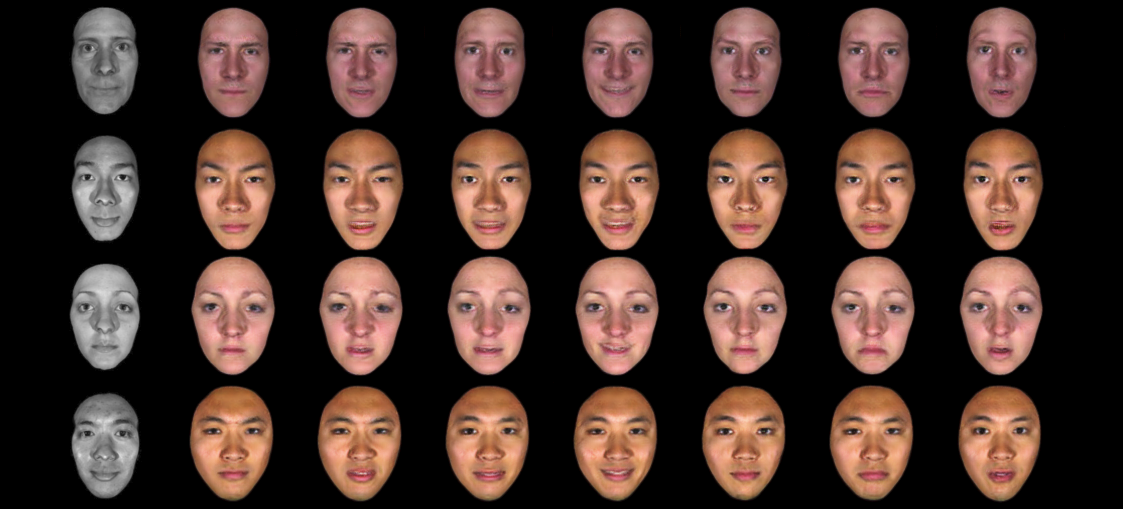}
   \caption{}
\end{subfigure}

\begin{subfigure}[b]{0.46\textwidth}
   \includegraphics[width=\linewidth]{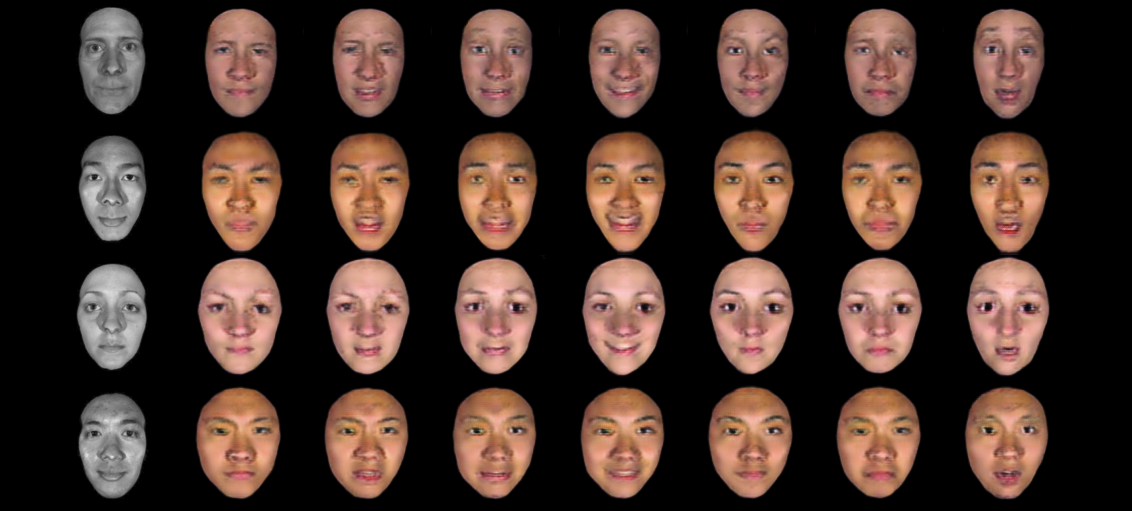}
   \caption{}
\end{subfigure}

\caption{Pose-normalized face attribute transfer results of (a) LSSL method compared with (b) SimGAN method \cite{shrivastava2017learning} on the BU-3DFE dataset \cite{yin20063d}. The input synthetic frontal face and real profile face are fed into our model to exhibit specified attribute. \textbf{Left to right}: input synthetic face and seven different attributes including \textit{angry}, \textit{disgusted}, \textit{fearful}, \textit{happiness}, \textit{neutral}, \textit{sadness} and \textit{surprised}, respectively.}
\label{fig:7}
\end{figure}

\paragraph{Qualitative evaluation} As shown in Fig. \ref{fig:6}, our facial attribute transfer test results (unseen images during the training step) are more visually pleasing compared to recent baselines including IcGAN \cite{perarnau2016invertible} and CycleGAN. \cite{zhu2017unpaired}. We believe that our proposed losses (parsing loss and identity losses) help to preserve the face image details and identity. IcGAN even fails to generate subjects with desired attributes, while our proposed method could learn attribute invariant features applicable to synthesize multiple images with desired attributes. In addition, to evaluate the proposed pose normalization method, the face attribute transfer results of our proposed method have been compared with the SimGAN method \cite{shrivastava2017learning} on the BU-3DFE dataset \cite{yin20063d} (see Fig. \ref{fig:7}). 

\begin{figure}[t]
\centering
\includegraphics[height=6.5cm, width=8.6cm]{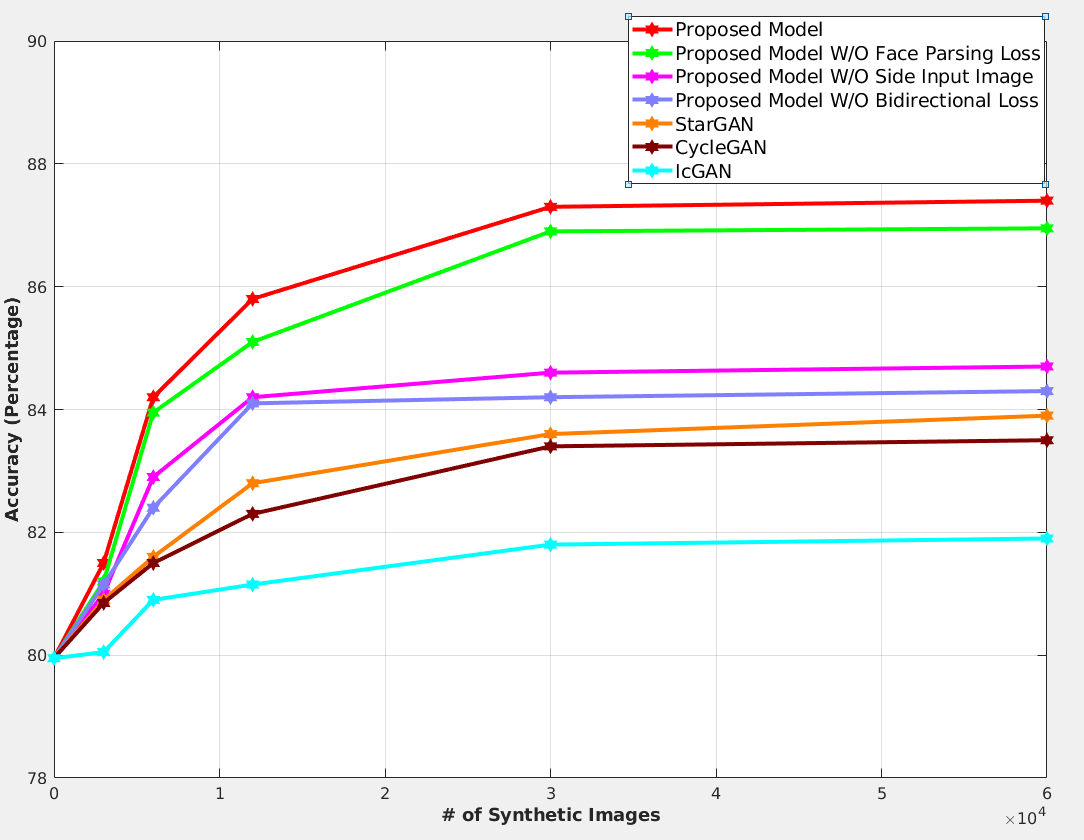}
  \caption{Impact of the amount of training synthetic images on performance in terms of expression recognition accuracy.}
\label{fig:8}
\end{figure}

\paragraph{Quantitative Evaluation} 
To conduct the quantitative analysis, we apply LSSL to data augmentation for facial expression recognition. We augment real images from Oulu-CASIA VIS dataset with the synthetic expression images generated by LSSL as well as its variants and then compare with other methods to train an expression classifier. The purpose of this experiment is to introduce more variability and enrich the dataset further, in order to improve the expression recognition performance. In particular, from each of the six expression category, we generate 0.5K, 1K, 2K, 5K and 10K images, respectively. As shown in Fig. \ref{fig:8}, when the number of synthetic images is increased to 30K, the accuracy is improved drastically, reaching to 87.40\%. The performance starts to become saturated when more images (60K) are used. We achieved a higher recognition accuracy value using the images generated from LSSL than other CNN-based methods including  popular generative model, StarGAN \cite{choi2018stargan} (see Table \ref{aug_synthesis}). This suggests that our model has learned to generate more realistic facial images controlled by the expression category. In addition, we evaluate the sensitivity of the results for different components of LSSL method (face parsing loss, bidirectional loss and side conditional image, respectively). We observe that our LSSL method trained with each of the proposed loss terms yields a notable performance gain in facial expression recognition.
\begin{table}[h]
\caption{Performance comparison of expression recognition accuracy between the proposed method and other state-of-the-art methods.}
\label{aug_synthesis}
\begin{center}
\begin{tabular}{|c||c|c|}
\hline
Method & Accuracy \\
\hline \hline
HOG 3D \cite{klaser2008spatio}& 70.63\%  \\
AdaLBP \cite{zhao2011facial}& 73.54\% \\
Atlases \cite{guo2012dynamic} & 75.52\% \\
STM-ExpLet \cite{liu2014learning} & 74.59\% \\
DTAGN \cite{jung2015deep} & 81.46\% \\
StarGAN \cite{choi2018stargan} & 83.90\% \\
\hline
\textbf{LSSL W/O Side Input} & \textbf{84.70\%} \\
\textbf{LSSL W/O Bidirectional Loss} & \textbf{84.30\%} \\
\textbf{LSSL W/O Face Parsing Loss} & \textbf{86.95\%} \\
\textbf{LSSL} & \textbf{87.40\%}\\
\hline
\end{tabular}
\end{center}
\end{table}

\begin{figure}
\centering
\begin{subfigure}[a]{0.46\textwidth}
   \includegraphics[width=\linewidth]{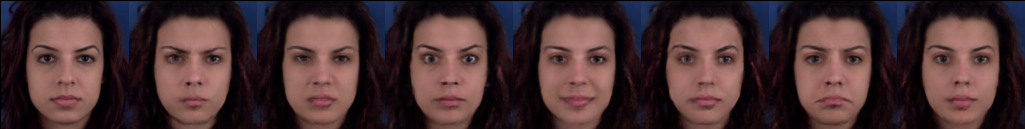}
   \caption{}
\end{subfigure}

\begin{subfigure}[b]{0.46\textwidth}
   \includegraphics[width=\linewidth]{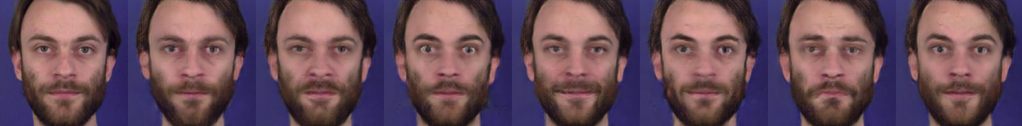}
   \caption{}
\end{subfigure}

\caption{Facial attribute transfer results from our proposed method for (a) subject 1 and (b) subject 20, respectively. The input face images are manipulated to exhibit desired attribute. \textbf{Left to right}: input \textit{neutral} face and seven different attributes including \textit{anger}, \textit{disgust}, \textit{fear}, \textit{happiness}, \textit{neutral}, \textit{sadness} and \textit{surprise}, respectively.}
\label{fig:9}
\end{figure}

Moreover, we evaluate the performance of LSSL on the MUG facial expression
dataset \cite{aifanti2010mug} using the video frames of the peak expressions. Fig. \ref{fig:9} shows sample facial attribute transfer results on the MUG facial dataset \cite{aifanti2010mug}. It should be noted that the MUG facial expression dataset are only available to authorized users. We only have permission from few subjects including 1 and 20 for using their photos in our paper. In Table \ref{MUG1}, we report the results of average accuracy of a facial expression on synthesized images. We trained a facial expression classifier with $\left ( 90\%/10\% \right )$ splitting for training and test sets using a ResNet-50 \cite{he2016deep}, resulting in a near-perfect accuracy of $90.42\%$. We then trained each of baseline models including CycleGAN, IcGAN and StarGAN using the same training set and performed image-to-image translation on the same test set. Finally, we classified the expression of these generated images using the above-mentioned classifier. As can be seen in Table \ref{MUG1}, our model achieves the highest classification accuracy (close to real image), demonstrating that our model could generate the most realistic expressions among all the methods compared.

\begin{table}
\caption{Performance comparison on the MUG dataset in terms of average classification
accuracy. }
\label{MUG1}
\begin{center}
\begin{tabular}{|c||c|c|}
\hline
Method & Accuracy \\
\hline \hline
Real Test Set  & 90.42\% \\
CycleGAN \cite{zhu2017unpaired} & 84.40\% \\
IcGAN \cite{perarnau2016invertible} & 80.32\% \\
StarGAN \cite{choi2018stargan}& 85.15\% \\
\textbf{LSSL W/O Face Parsing Loss}& \textbf{89.91\%} \\
\textbf{LSSL} & \textbf{90.35\%} \\
\hline
\end{tabular}
\end{center}
\end{table}

\paragraph{Pose Normalization Analysis} Using BU-3DFE dataset \cite{yin20063d}, we have designed subject-independent experimental setup. We performed 5-fold cross validation using 100 subjects. Training data includes images of 80 (frontal face) subjects, while test data includes images of 20 subjects with varying poses. We use VGG-Face model \cite{parkhi2015deep}, which is pretrained on the (RaFD) \cite{langner2010presentation} and then we further fine-tune it on the frontal face images from BU-3DFE dataset. It can be observed from Table \ref{frontal_cnn} that pose normalization helps to improve expression recognition performance of the non-frontal faces (ranging from 15 to 45 degrees in 15 degrees steps). Having said that, adding realism to simulated face images helps to bring additional gains in terms of expression recognition accuracy. In particular, our method outperforms two recent works, \cite{lai2018emotion,zhang2018joint} that addressed pose normalization task. Our proposed losses (parsing loss and identity losses) facilitates the synthesized frontal face images to preserve much detail of face characteristics (e.g. expression and identity).
 \begin{table}[h]
		\caption{Recognition accuracies on normalized face images at different pose angles.}
		\begin{center}
		\resizebox{0.48\textwidth}{!}{
   \begin{tabular}{|c|c|c|c|c|}
   \hline
Method & \textbf{$\pm 15$} & \textbf{$\pm 30$} & \textbf{$\pm 45$}  \\ \hline
Real Profile Face Images         & 70.15\%        & 66.50\%        & 58.90\%               \\
Simulated Frontal Face Images         & 70.91\%        & 65.90\%        & 59.30\%               \\
CycleGAN         & 71.60\%        & 67.32\%        & 61.50\%               \\
\cite{lai2018emotion} & 71.45\%        & 67.60\%        & 61.95\%               \\
\cite{zhang2018joint}  & 71.72\%        & 67.65\%        & 62.10\%               \\
\textbf{LSSL W/O Face Parsing Loss}         & \textbf{72.10\%}        & \textbf{68.52\%}        & \textbf{63.35\%}             \\
\textbf{LSSL}         & \textbf{72.70\%}        & \textbf{69.10\%}        & \textbf{64.05\%}             \\
 \hline  
\end{tabular}}
    \end{center}
    \label{frontal_cnn}
\end{table} 

\subsection{Visualizing Representation}
\begin{figure}[h]
    \begin{subfigure}{\textwidth}
    \includegraphics[height=1.3cm, width=8.9cm]{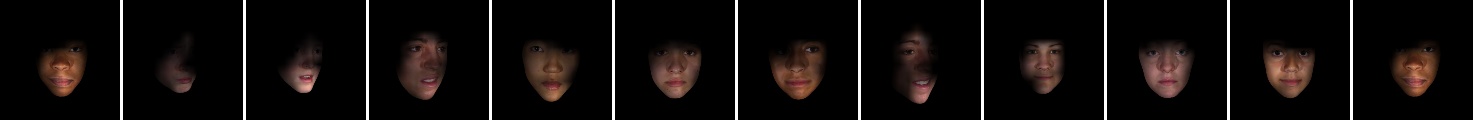}
    \label{fig:doc1}
    \end{subfigure}
    \bigskip
    \begin{subfigure}{\textwidth}
    \includegraphics[height=1.3cm,width=8.9cm]{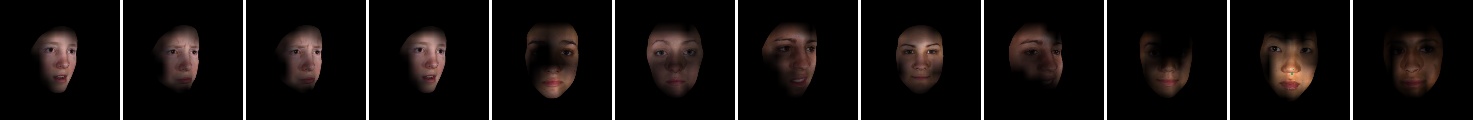}
    \label{fig:doc2}
    \end{subfigure}
    \bigskip
    \begin{subfigure}{\textwidth}
    \includegraphics[height=1.3cm,width=8.9cm]{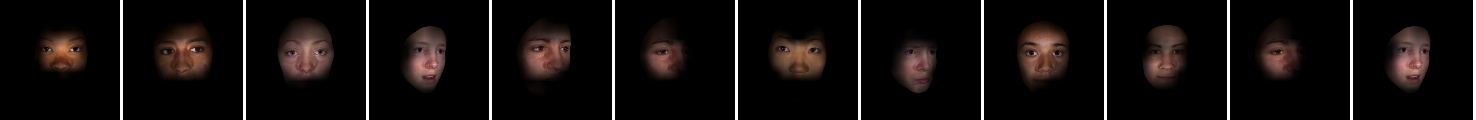}
    \label{fig:doc3}
    \end{subfigure}
        \bigskip
    \begin{subfigure}{\textwidth}
    \includegraphics[height=1.3cm,width=8.9cm]{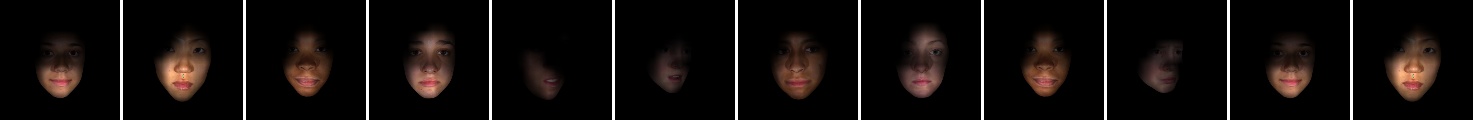}
    \label{fig:doc3}
    \end{subfigure}
\caption{Visualization of some hidden units in the encoder of LSSL trained on the BU-3DFE dataset \cite{yin20063d}. We highlight regions of face images that a particular convolutional hidden unit maximally activates on.}
\label{fig:10}
\end{figure}

Fig. \ref{fig:10} visualizes some activations of hidden units in the fifth layer of an encoder (the first component of the generator). Although all units are not semantic, but these visualizations indicate that the network learns to identity the most informative visual cues from the face regions.

\subsection{Training Losses Additional Qualitative Results}
\begin{figure*}[t]
\begin{subfigure}{.5\textwidth}
  \centering
  \includegraphics[width=9cm, height=6cm]{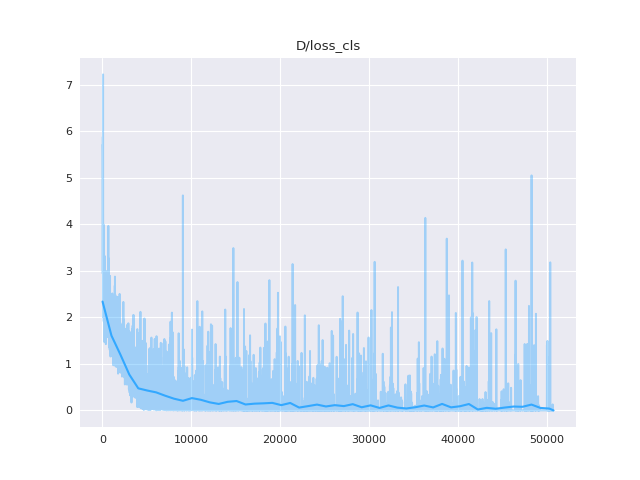}
  \caption{}
  \label{fig:sfig1}
\end{subfigure}%
\begin{subfigure}{.5\textwidth}
  \centering
  \includegraphics[width=9cm, height=6cm]{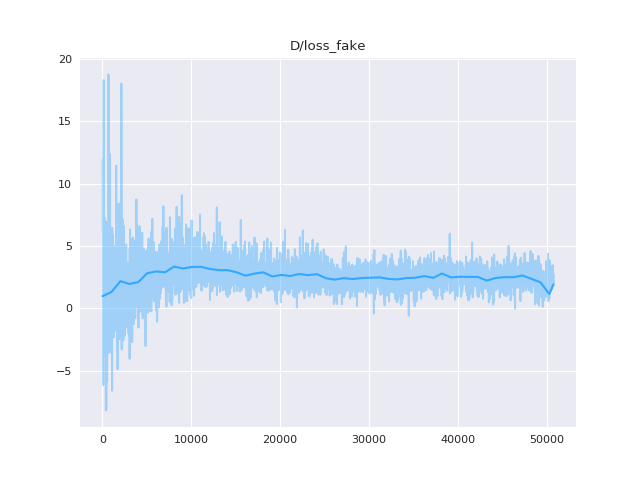}
  \caption{}
  \label{fig:sfig2}
\end{subfigure}
\begin{subfigure}{.5\textwidth}
  \centering
  \includegraphics[width=9cm, height=6cm]{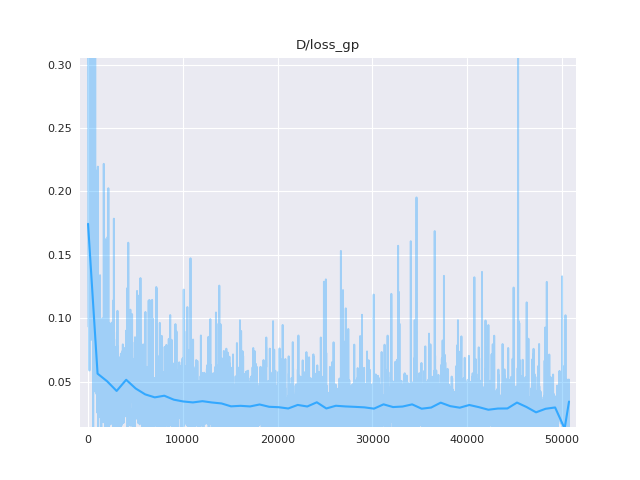}
  \caption{}
  \label{fig:sfig2}
\end{subfigure}
\begin{subfigure}{.5\textwidth}
  \centering
  \includegraphics[width=9cm, height=6cm]{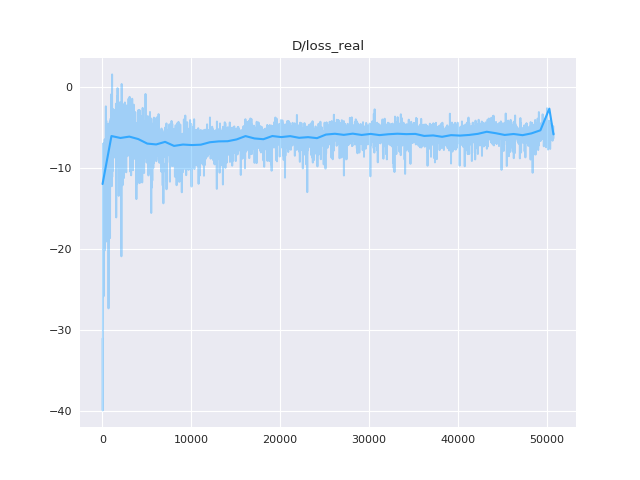}
  \caption{}
  \label{fig:sfig2}
\end{subfigure}
\caption{The training losses for the network's discriminator on the RaFD dataset \cite{langner2010presentation}. (a) An attribute classification loss of real images, (b) the discriminator loss for the pair of fake generated images, (c) discriminator gradient penalty loss and (d) discriminator loss for the pair of real images, respectively.}
\label{fig:11}
\end{figure*}

\begin{figure*}[h]
\centering
\includegraphics[height=22cm, width=16cm]{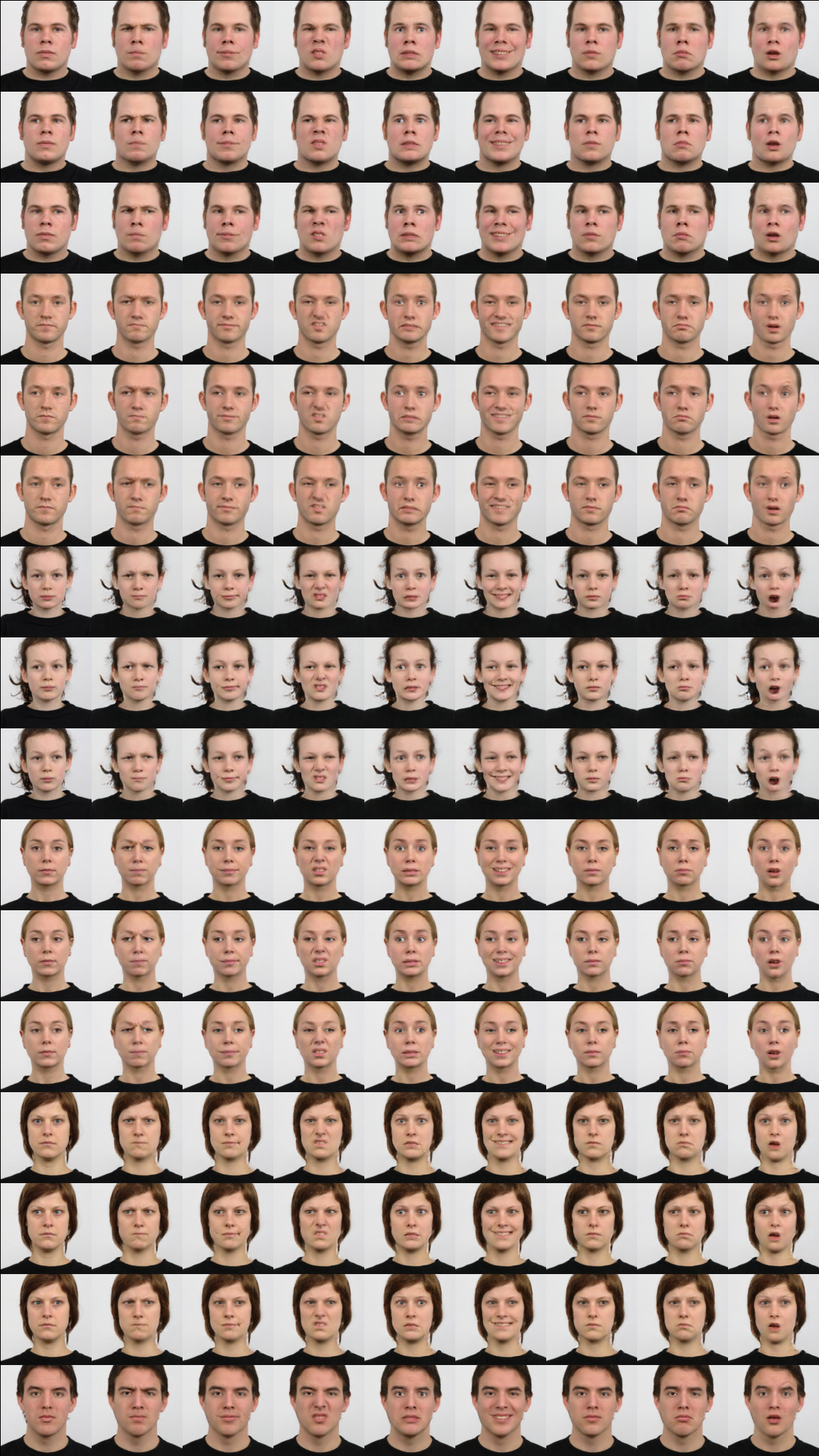}
  \caption{Facial expression synthesis on the Radboud Faces Database (RaFD).
  \textbf{Left to right}: input \textit{neutral} face and synthesis results of all eight emotion classes including \textit{angry}, \textit{contemptuous}, \textit{disgusted}, \textit{fearful}, \textit{happiness}, \textit{neutral}, \textit{sadness} and \textit{surprised}, respectively.}
\label{fig:12}
\end{figure*}

Fig. \ref{fig:11} shows the training losses of the proposed attribute guided face image synthesis model for the discriminator. Here, we use the face landmark heatmap as the side conditional image. The face landmark heatmap contains 2D Gaussians centered at the landmarks' locations, which are then concatenated with the input image to synthesize different facial expressions on the RaFD dataset \cite{langner2010presentation}. In addition, the target attribute label is spatially replicated and concatenated with the latent feature. Results in Fig. \ref{fig:11} are for 100 epochs, 50,000 iterations of training on the RaFD dataset. Moreover, Fig. \ref{fig:12} shows additional images generated by LSSL.
\section{Conclusion}
In this work, we introduced LSSL, a model for multi-domain image-to-image translation
applied to the task of face image synthesis. We present attribute guided face image generation to transform a given image to various target domains controlled by desired attributes. We argue that learning image-to-image translation between image domains requires a proper modeling the shared latent representation across image domains.
Additionally, we proposed face parsing loss and identity loss to preserve much detail of face characteristics (e.g. identity). More importantly, we seek to add realism to the synthetic images while preserving the face pose angle. We also demonstrate that the synthetic images generated by our method can be used for data augmentation to enhance facial expression classifier's performance. We reported promising results on the task of domain adaptation by adding the realism to the simulated faces. We showed that by leveraging the synthetic face images as a form of data augmentation, we can achieve significantly higher average accuracy compared with the state-of-the-art result.
%


\bibliographystyle{model2-names}
\bibliography{egbib.bib}
\end{document}